\documentclass{article} 
\usepackage{collas2026_conference,times}
\usepackage{easyReview}

\usepackage{amsfonts,amsmath,amssymb}
\usepackage{subcaption}
\usepackage{algorithm}
\usepackage{algpseudocode}
\usepackage{wrapfig}
\usepackage{float}
\usepackage{algpseudocode}
\usepackage{tablefootnote}
\usepackage{booktabs}
\usepackage{tabularx}

\providecommand{\sct}[1]{{\sc \texttt{#1}}}

\usepackage{amsmath,amsfonts,bm}

\def\eqref#1{equation~\ref{#1}}

\def\1{\bm{1}}

\DeclareMathAlphabet{\mathsfit}{\encodingdefault}{\sfdefault}{m}{sl}
\SetMathAlphabet{\mathsfit}{bold}{\encodingdefault}{\sfdefault}{bx}{n}

\DeclareMathOperator*{\argmax}{arg\,max}

\usepackage{hyperref}
\hypersetup{
    colorlinks=true,
    linkcolor=red,
    filecolor=magenta,
    urlcolor=blue,
    citecolor=purple,
    pdftitle={Overleaf Example},
    pdfpagemode=FullScreen,
    }

\title{SHARP: Sleep-based Hierarchical Accelerated Replay for Long Range Non-Stationary Temporal Pattern Recognition}

\author{
Jayanta Dey$^{1}$ \quad
Shikhar Srivastava$^{2}$ \quad
Itamar Lerner$^{1}$ \quad
Christopher Kanan$^{2}$ \quad
Dhireesha Kudithipudi$^{1}$ \\
\\
$^{1}$University of Texas at San Antonio, USA \\
$^{2}$ University of Rochester, USA \\
\\
}

\collasfinalcopy 

\begin{document}

\maketitle


\begingroup
\renewcommand\thefootnote{}
\footnote{Implementation of \sct{SHARP}\ is available at \url{https://github.com/jdey4/sharp}. Email: deyjayanta76@gmail.com}
\addtocounter{footnote}{-1}
\endgroup

\begin{abstract}
Learning long-range non-stationary temporal patterns remains a core challenge for modern sequence models, particularly in strict streaming settings. In these settings, data arrive sequentially and must be processed in a single pass without simultaneously revisiting past observations. Standard architectures, including recurrent neural networks and transformers, are constrained by either truncated backpropagation through time horizon or explicit input window length for long range credit assignment. To address these limitations, we propose \sct{SHARP} (\textit{Sleep-based Hierarchical Accelerated Replay}), a framework that decomposes temporal learning into two complementary components: a memory module that accumulates a structured history of past inputs, and a pattern-recognition module that operates over this memory. This separation enables resource- and compute-efficient adaptation to non-stationary dynamics by eliminating the need for backpropagation through time across many steps for long-range credit assignment. Inspired by the accelerated replay observed in rodents during slow-wave sleep, \sct{SHARP} incorporates offline (sleep) phases in which temporally structured memory traces are replayed in an accelerated form and integrated into higher-level memory representations, improving long-range context retention. Through controlled simulations and ablation studies, we characterize the key properties of the proposed framework. In benchmark datasets such as \textsc{text8} and \textsc{PG-19}, we demonstrate that \sct{SHARP} improves over recurrent baselines by retaining next-token predictive performance on previously seen data while continuing to learn from the current stream and generalizing to future unseen data. These gains are enabled by its hierarchical structure, which yields an exponentially increasing effective temporal context with only linear-time computational cost.
\end{abstract}
\section{Introduction}

In many real-world settings, observations arrive sequentially without the possibility of revisiting past data. Learning algorithms must therefore continually integrate new information while preserving the structure of prior experience \citep{harunsiesta}. This imposes a strict constraint: learning must proceed online, with limited opportunity for long-horizon credit assignment. The challenge is further exacerbated under distribution shift, where the underlying data-generating process evolves over time.

From a modeling perspective, continual learning under streaming constraints can be naturally formulated as a sequential learning problem. To generalize under these constraints, a system must retain information about past inputs even after they are no longer directly accessible. 
Classical sequence models such as recurrent neural networks (\sct{RNNs}) and long short-term memory networks (\sct{LSTMs}) attempt to encode memory within recurrent dynamics. However, their effective memory is governed by backpropagation through time (BPTT), which limits credit assignment to a finite temporal horizon and introduces numerical instabilities such as vanishing and exploding gradients. Although recurrent models have a theoretically unbounded context memory, in practice their memory is lossy: information dissipates, interferes, or becomes entangled over time, restricting the reliable capture of long-range temporal structure \citep{bengio1994learning, hochreiter1997long}. A common consequence of the limited temporal horizon is the degradation of previously acquired knowledge as new information is incorporated, known as catastrophic forgetting \citep{mccloskey1989catastrophic,mcclelland1995there,doan2021theoretical,vogelstein2025simple}. Indeed, one way to understand catastrophic forgetting is as a consequence of limited long-range credit assignment, which biases learning toward the current task and degrades generalization to past tasks.

Existing regularization-based approaches, developed to mitigate catastrophic forgetting, do not treat memory as an explicit structural component; instead memory in these models emerges implicitly through gradient-based optimization. For instance, a neural network trained sequentially on multiple tasks tends to overwrite previously learned representations unless additional mechanisms, such as Elastic Weight Consolidation (\sct{EWC}) \citep{kirkpatrick2017overcoming} or Learning without Forgetting (\sct{LwF}) \citep{li2017learning}, are introduced to preserve past knowledge in the weights. Alternatively, replay-based approaches maintain external buffers to revisit past samples \citep{shin2017continual, chaudhry2019tiny, tolias_architecture, buzzega2020dark,channappayya2023augmented}. In these models, replay often serves a composite role, bundling together explicit data storage, old-task rehearsal, and the refinement of predictive models through supervised updates \citep{lopez2017gradient, rebuffi2017icarl, shin2017continual, chaudhry2018efficient, rolnick2019experience, buzzega2020dark}. Despite their differences, the above strategies treat memory either as constrained weight plasticity or as stored raw data, rather than as a structured dynamical system. In general, an organized memory system is essential for generalization in a sequential and continually evolving environment \citep{dorovatas2026modular}.

Biological systems appear to circumvent the above limitations through structured memory organization that takes places, at least partially, during sleep \citep{o2014complementary, kumaran2016learning, lutz2026sleep}. In particular, evidence from rodents suggests that during slow-wave sleep (SWS), sequential memory of previously encoded experiences are reactivated in the hippocampus in a shorter timescale than originally experienced, essentially implementing accelerated (or ``time-compressed'') replay. One recent theory, the temporal scaffolding hypothesis (TSH) posits that such accelerated replay has a functional role, enabling the consolidation of long-range associations that are difficult to form during online experience alone \citep{Lerner2017, lerner2018individual, lerner2019sleep, lerner2022sleep}.

\begin{figure}[!t]
    \vspace{-10pt}
    \centering
    \includegraphics[width=.65\textwidth]{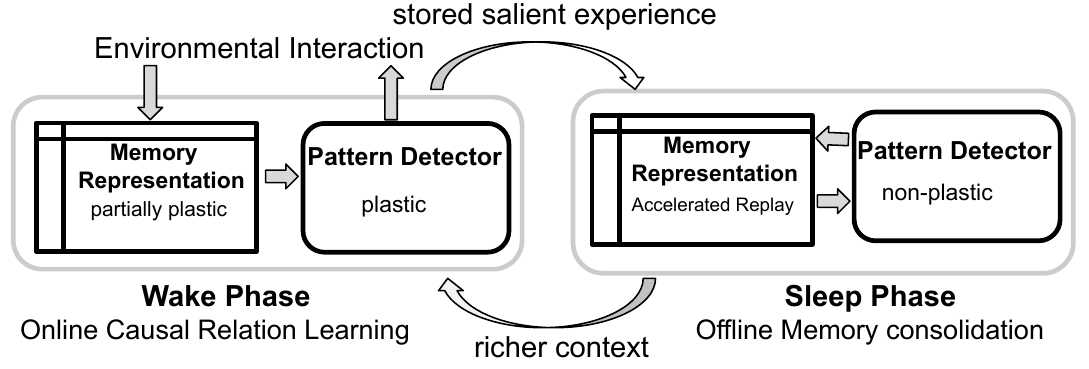}
 \caption{\textbf{Conceptual overview of slow-wave sleep-based temporal learning.} During wake, environmental interaction drives causal relation learning by updating plastic memory and pattern modules. Salient experiences are tagged and later replayed during sleep, where accelerated replay provides richer temporal context for memory consolidation. 
 }
\label{fig:overview}
\vspace{-10pt}
\end{figure}

\begin{figure}[!t]
    \vspace{-5pt}
    \centering
    \includegraphics[width=.85\textwidth]{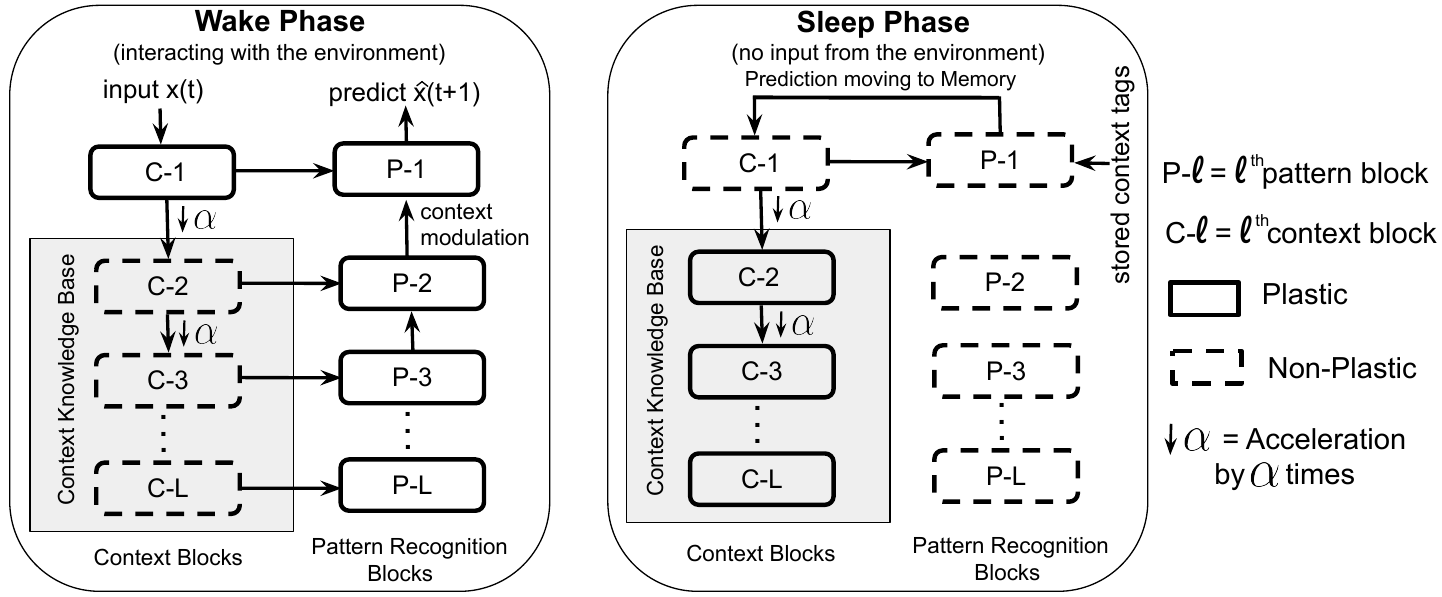}
 \caption{\textbf{Sleep-based hierarchical accelerated replay framework.} \textit{Left (Wake phase):} The context knowledge base (upper context blocks) remains non-plastic while the model interacts with the environment. Lower context-block memories are progressively accelerated into their immediately higher blocks. \textit{Right (Sleep phase (SWS)):} The context knowledge base is updated offline, while C-1 and P-1 stop receiving environmental input and feedback, respectively and replays its tagged wake experiences to the context knowledge base. 
 }
\label{fig:model}
\vspace{-10pt}
\end{figure}

Informed by TSH, the current work presents a new learning framework, SHARP (Sleep-based Hierarchical Accelerated Replay), that learns to detect temporal patterns using a “wake” phase and a ``sleep'' phase (Figure \ref{fig:overview}). The sleep phase incorporates an abstracted approximation of accelerated replay (explained in detail in Section \ref{sec:acceleration}) that allows the system to traverse longer temporal contexts during consolidation than would be feasible during online learning, thus improving memory retention. Replay during sleep is restricted to unsupervised memory consolidation to allow better future memory retention and prediction, while the learning of causal and predictive relationships is driven exclusively by wake-time interactions with the environment. These two processes are accomplished within two separate modules: (i) a hierarchical memory module that accumulates experience without credit assignment from prediction and (ii) a hierarchical pattern-recognition module that operates over this memory to perform prediction (Figure \ref{fig:model}). This separation avoids confounding memory storage with credit assignment and provides a stable substrate for pattern learning.

Unlike modern state space models (\sct{SSM}s) \citep{gu2021efficiently, gu2023mamba}, where memory is encoded within a single dynamical system, our framework explicitly learns and organizes memory representations hierarchically across multiple layers (see Figure \ref{fig:model}). During the wake phase, only the lowest layer remains plastic, accumulating recent experiences. Salient events are selectively tagged \citep{yang2024selection} and later consolidated into higher layers during offline sleep phases through accelerated sequential replay. The higher layers thus form a stable context knowledge base, while lower layers capture rapidly changing dynamics. This hierarchical consolidation expands the effective temporal context available to downstream pattern-recognition modules in a familiar environment, while keeping most parameters frozen during online interaction. In this paper, we demonstrate that accelerated replay effectively extends the model’s usable context window, enabling it to capture long-range dependencies without requiring long-horizon BPTT. In what follows, we describe the problem setting, formalize the desired properties of memory and acceleration, describe the model architecture and learning dynamics, and evaluate an instantiation of the proposed framework through controlled simulations and two benchmark datasets.

\section{Technical Background}

\subsection{Problem Setting}
\label{sec:problem}
Let $\{X_1, X_2, \cdots, X_t \}$ be a stochastic process or a sequence of random variables where each variable takes a value from some finite set $\mathcal{A} = \{a_1, a_2, \cdots, a_K\} \subseteq \mathbb{R}^D$. The variable $X_t$ represents the state of the process at time $t$, governed by a set of underlying state transition probability laws $\mathcal{P} = \{\mathbf{p}_1, \mathbf{p}_2, \cdots, \mathbf{p}_t\}$. Each $\mathbf{p}_t \in \mathcal{P}$ defines the probability of transitioning to any state $a_i \in \mathcal{A}, \forall i=1,\cdots,K$ given all previous states up to time $(t-1)$, that is, 

\begin{equation}
    \mathbf{p}_t = [p_t(a_i)]_{i=1}^K = [P(X_t=a_i|X_1,\cdots,X_{t-1})]_{i=1}^K.
\end{equation}

Given a sequence of states evolving according to an unknown transition rule $\mathcal{P}$, a sequential learner $f: \mathcal{A}^T \rightarrow [0, 1]^K$, having access to an input window of past $T$ states, estimates $\mathbf{p}_t$ :

\begin{equation}
    \hat{\mathbf{p}}_t = [\hat{p}_n(a_i)]_{i=1}^K =  f(x_{t-T}, \cdots, x_{t-1} ),
\end{equation}

where $x_t$ is the value of the random variable $X_t$ at time $t$. The state at time $t$ is estimated as the $\argmax$ of $\hat{\mathbf{p}}_t$:

\begin{equation}
    \hat{x}_t = \argmax_{a_i \in \mathcal{A}} \hat{p}_t(a_i),  \forall i=1,\cdots,K.
\end{equation}
If the transition probability between states, $\mathbf{p}_t$, depends on more past states than those captured within the input window, then the learner must integrate an internal mechanism to retain the memory of the previous states. The estimation accuracy of the next sample depends on how effectively the model retains and utilizes past information. Unlike traditional training setups, we adopt a one-pass learning regime in which the model observes training samples in an online streaming manner and cannot optimize for multiple parallel sequence segments simultaneously.

\paragraph{Memory encoding desiderata}
While prior work has explored various forms of memory, including associative memory \citep{hopfield1982neural, kanerva1988sparse} and sparse or quantized representations \citep{olshausen1997sparse, van2017neural, razavi2019generating}, these approaches primarily emphasize storage or compression. In contrast, we focus on the \emph{dynamics} of memory, i.e., how information about past inputs is continuously maintained and updated over time. To support stable downstream processing, we seek continuous latent states that preserve similarity structure across inputs, enabling pattern-recognition modules to operate on temporally coherent signals without explicit access to past samples \citep{ba2016using}.

In our framework, a pattern-recognition map $f(\cdot)$ operates on a dynamic memory encoding $m(\cdot, \cdot)$ to produce the current state transition probability, i.e., $\hat{\mathbf{p}}_t = f\big(m(x_{t-1}, h_{t-1})\big),$
where $h_{t-1}$ denotes the previous memory state. We now describe desirable properties of the memory encoding $m(\cdot, \cdot)$. 

Let $\mathcal{S}= \{S_1, S_2, \cdots\}$ denote the set of all length-$s$ sequences over the elements of $\mathcal{A}$, equipped with a metric $d_{\mathcal{S}}$. For a sequence $S_i \in \mathcal{S}$, let $h_i = m(S_i, h'_i) \in \mathcal{H} \subseteq \mathbb{R}^P$ denote the encoding after observing the $s$-th sample, where $(\mathcal{H}, d_{\mathcal{H}})$ is a metric space and $h'_i$ is the previous state. Ideally, we seek memory encodings that approximately preserve similarity structure across sequences, such that sequences that are close under $d_{\mathcal{S}}$ map to nearby representations under $d_{\mathcal{H}}$, up to bounded distortion. This can be expressed as:
\begin{equation}\label{eq:mem_property}
\frac{1}{C} d_{\mathcal{S}}(S_i, S_j)  \leq d_{\mathcal{H}}(h_i, h_j)  \leq C \, d_{\mathcal{S}}(S_i, S_j),
\end{equation}
for some constant $C > 0$. This perspective ensures that similar sequences map to nearby representations while remaining distinguishable. We use the notions of \emph{memory capacity} as the size of the largest subset of $\mathcal{S}$ whose elements can remain distinguishable under the above representation, and \emph{memory span} as the temporal extent or sequence length $s$ over which information is retained. These terms serve as conceptual tools for characterizing memory encodings.

\subsection{Acceleration}
\label{sec:acceleration}

We use acceleration as a computational abstraction of compressed-time biological replay. In SHARP, this is implemented through temporal downsampling over learned memory states as information passes from one layer to the next one: Each higher layer receives a downsampled sequence of the lower-layer’s states, so each higher-level transition summarizes a longer span of environment-level experience. During wake, the environment determines the pace of incoming inputs; therefore, downsampling makes higher-layer states evolve more slowly (they have to “wait” to the environmental inputs to proceed before updates can occur). However,  during sleep, replay is not paced by environmental input, allowing higher layers to traverse memory states faster than online experience; (Figure \ref{fig:model}).
Therefore, the combination of downsampled states, each containing information on several timesteps from the previous layer,  and replay of these states free from the environmental pace, effectively creates acceleration of inputs from one layer to the next.

\begin{wrapfigure}[13]{r}{0.6\textwidth}
    \vspace{-15pt}
    \centering
    \includegraphics[width=0.58\textwidth]{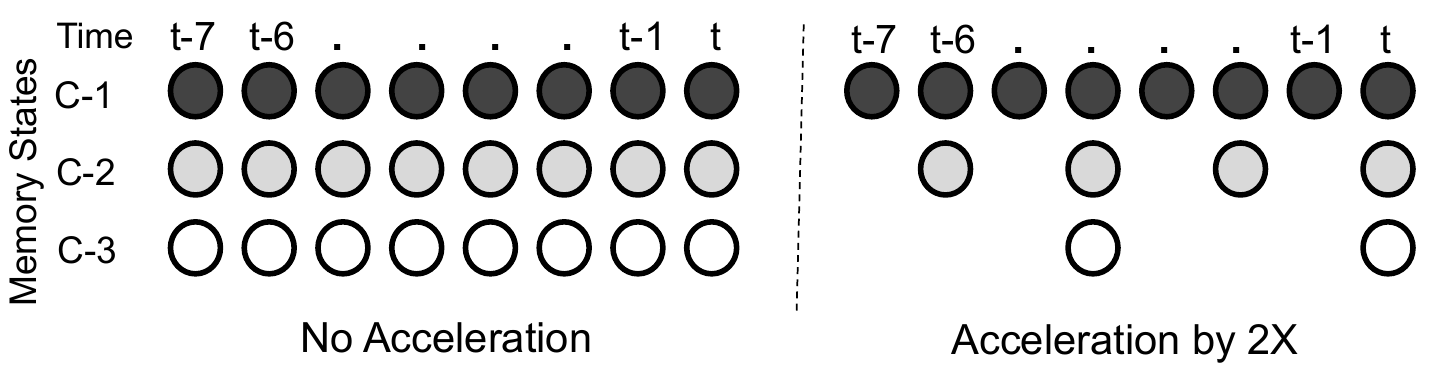}
    \caption{\textbf{Wake–sleep temporal scaling in \sct{SHARP} (see Figure \ref{fig:model}).} During wake, higher layers update at slower rates (C-2 updates once every two environment-level steps). During sleep, C-2 input states are generated sequentially from C-1. Thus generating four C-2 level input states corresponds to traversing eight environment-level inputs in four steps.}
    \label{fig:concept}
\end{wrapfigure}


We define acceleration as passing only every $\alpha$-th lower-level memory state to the next higher level. 
Suppose a lower-level memory state $h_t$ summarizes the most recent $s$ inputs, i.e., the window $\{x_{t-s+1}, \ldots, x_t\}$. 
If the higher level receives the subsequence $\{h_1, h_{1+\alpha}, h_{1+2\alpha}, \ldots\}$, then it processes fewer states while each received state summarizes multiple environment-level inputs. 
When the memory encoding is sufficiently lossless, this downsampled sequence still preserves the relevant temporal structure but allows the higher level to traverse a longer environment-level history per update. We call $\alpha$ the acceleration factor. Figure \ref{fig:concept} shows the update schedule of a $3$ layer hierarchical acceleration by $2\times$.



\subsection{Framework Description}
Figure~\ref{fig:model} illustrates the general architecture of \sct{SHARP}. We will describe an instance of the above generalized framework in detail in Section \ref{sec:sim}. As depicted in the figure, the model consists of a total of $L$ hierarchical blocks, each comprising a context block and a corresponding pattern-recognition block. The context blocks are organized bottom-up and the pattern recognition blocks are arranged top-down. 
The memory contents in the context blocks are not credit-assigned from the prediction objective, and hence gradients do not flow from the pattern recognition blocks back to the memory blocks. Memory in each context block is accelerated in the form of down-sampling by a factor of $\alpha$, as explained above, when transferred from a lower block to its immediate upper block. Note that this acceleration is possible because the memory operates without credit assignment. In practice, however, memory representations that rely on credit assignment may be imperfect. In particular, when the temporal credit assignment at the lowest level is constrained by a limited input horizon, the resulting memory states may not capture long-range dependencies. Under such conditions, acceleration can amplify these distortions, as higher-level blocks operate on accelerated representations derived from imperfect lower-level memory. This motivates the need for smooth, sufficiently lossless memory encodings that do not depend on credit assignment. Importantly, \sct{SHARP} induces a bootstrapping process through sleep-time replay: higher-level memory (e.g., C-2) starts noisy but improves through replay during sleep. This leads to better predictions in the next wake phase. The improved predictions then improve the quality of future replay (see Figure \ref{fig:generation}). Repeating this cycle progressively refines the overall representation.

During wake-time, the upper context blocks evolve on exponentially slower timescales compared to that of C-$1$: for every $\alpha$ updates of C-$1$, C-$2$ is updated once, and more generally, C-$\ell$ is updated once every $\alpha^{\ell-1}$ updates of C-$1$ (see Figure \ref{fig:concept}). Similarly, the learning rate for the upper pattern blocks is kept $\gamma \times$ slower than that of their immediate lower pattern block. This is done so that the upper pattern blocks do not overfit to transient lower-level fluctuations. This hierarchical separation of timescales prevents pattern-recognition modules from overfitting to transient input, instead encouraging them to operate over a broader temporal context, thereby improving generalization and reducing forgetting. Importantly, it also governs how upper blocks are trained during sleep. Once the model detaches from the environment, learning is no longer constrained by the pace of incoming data and does not require waiting for new observations to arrive. Consequently, upper blocks can be trained rapidly on accelerated memory traces generated by lower blocks. This sleep-time training optimizes the upper memory representation to better retain the trajectory of the current experience. In this work, \sct{SHARP} enters the sleep phase at a fixed interval. However, in essence, the sleep-mode could be triggered based on more real-world conditions encountered by biological systems, for example, following a period of inactivity due to diminished environmental inputs as occurring during nighttime.

\section{Mechanistic Simulation Studies of Context and Pattern Blocks}
\label{sec:sim}
\begin{wrapfigure}[16]{r}{0.52\textwidth}
    \vspace{-20pt}
    \centering
    \includegraphics[width=0.5\textwidth]{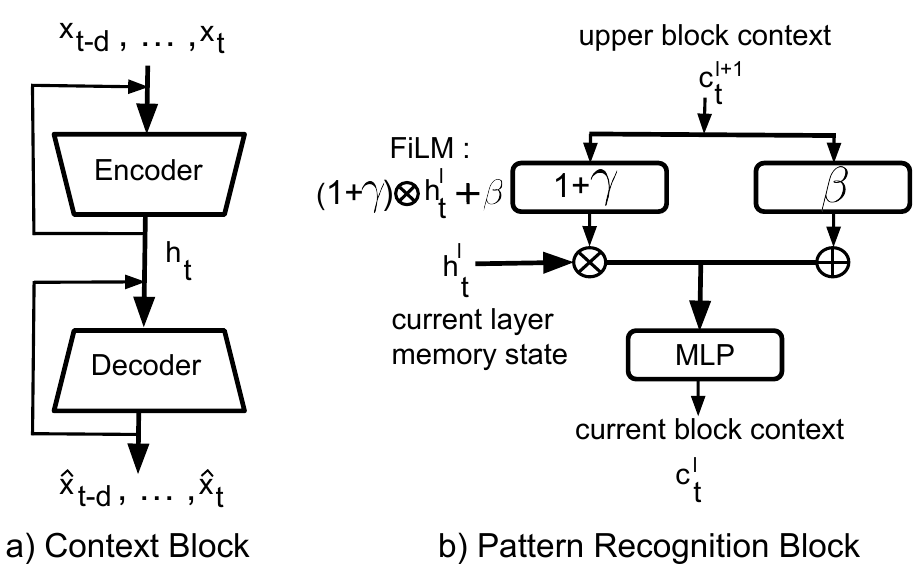}
    \caption{\textbf{Example context and pattern recognition blocks.}
    (a) Context encodes a window of $s$ inputs into $h_t$ and reconstructs without credit assignment.
    (b) Pattern block applies FiLM using context $c_t^{l+1}$ to modulate $h_t^{l}$ and produce $c_t^{l}$.}
    \label{fig:mem_pat}
\end{wrapfigure}

Before the detailed analyses, we summarize the instantiated architecture. \sct{SHARP} uses $L$ context blocks for hierarchical memory and $L$ pattern-recognition blocks for prediction. During wake, the lowest context block processes the stream while prediction losses update only pattern blocks; during sleep, tagged wake states are replayed offline to train higher context blocks through temporally downsampled memory traces.

\subsection{Architecture Instantiation}
Figure~\ref{fig:model} presents a general framework and in this work, we instantiate the context blocks using recurrent autoencoders. Specifically, we employ an RNN encoder–decoder architecture \citep{sutskever2014sequence}, as illustrated in Figure~\ref{fig:mem_pat}a, where the encoder maps an input window to a latent state and the decoder reconstructs the sequence from this representation during training. To promote continuity in the learned representations, the hidden state of each input window is initialized from the previous window while training. This encourages the memory to evolve smoothly over time and retain information across overlapping windows.

Figure \ref{fig:mem_pat}b illustrates the pattern recognition block used in this work. At each level, the upper-level context modulates the current-layer memory state via Feature-wise Linear Modulation (FiLM) \citep{perez2018film}. Specifically, the memory state undergoes feature-wise affine conditioning, where scaling and shifting parameters are generated from the upper-layer context. The modulated representation is then processed by a multilayer perceptron (MLP) to produce an updated context. This context is subsequently passed to the next lower layer or used for next-token prediction at the lowest layer. Below, we examine the empirical properties of each block.

\begin{figure}[!ht]
    \centering
    \vspace{-15pt}
    \includegraphics[width=0.75\textwidth]{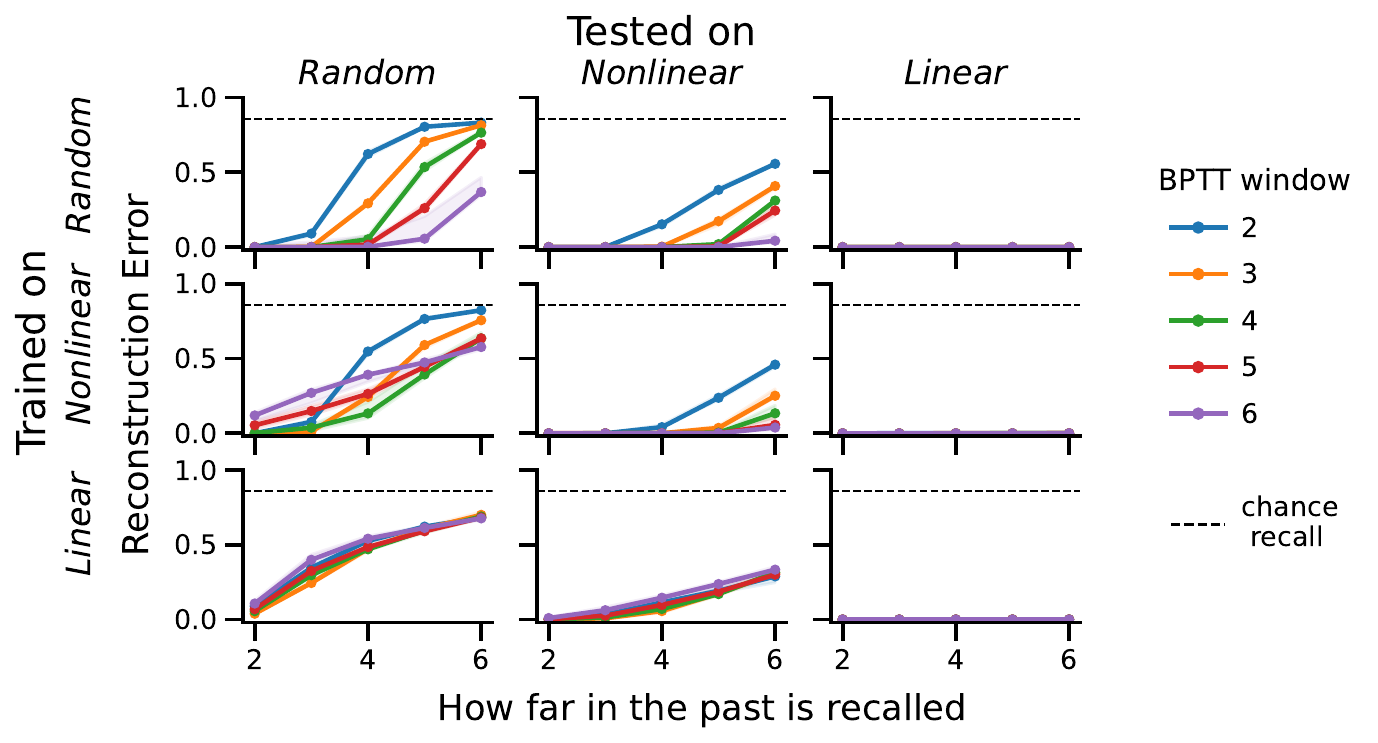}
 \caption{\textbf{Memory capabilities show carryover across regimes, with stronger transfer generally observed from harder (e.g., nonlinear) to easier sequences (e.g., linear).} Colors denote BPTT steps; shading shows interquartile range across $10$ runs; dashed line is chance performance.}

    \label{fig:mem_cap}
\end{figure}

\subsection{Simulation Environments}
We use three simulation environments to systematically probe different properties of the instantiated blocks: linear, nonlinear and random. See Appendix \ref{app:sim} for details. In terms of resource requirements and sequence compressibility, the difficulty of retaining the above three sequences can be ordered as:
\text{Random} \;>\; \text{Nonlinear} \;>\; \text{Linear}.

\begin{wrapfigure}[17]{r}{0.38\textwidth}
    \vspace{-15pt}
    \centering
    \includegraphics[width=0.34\textwidth]{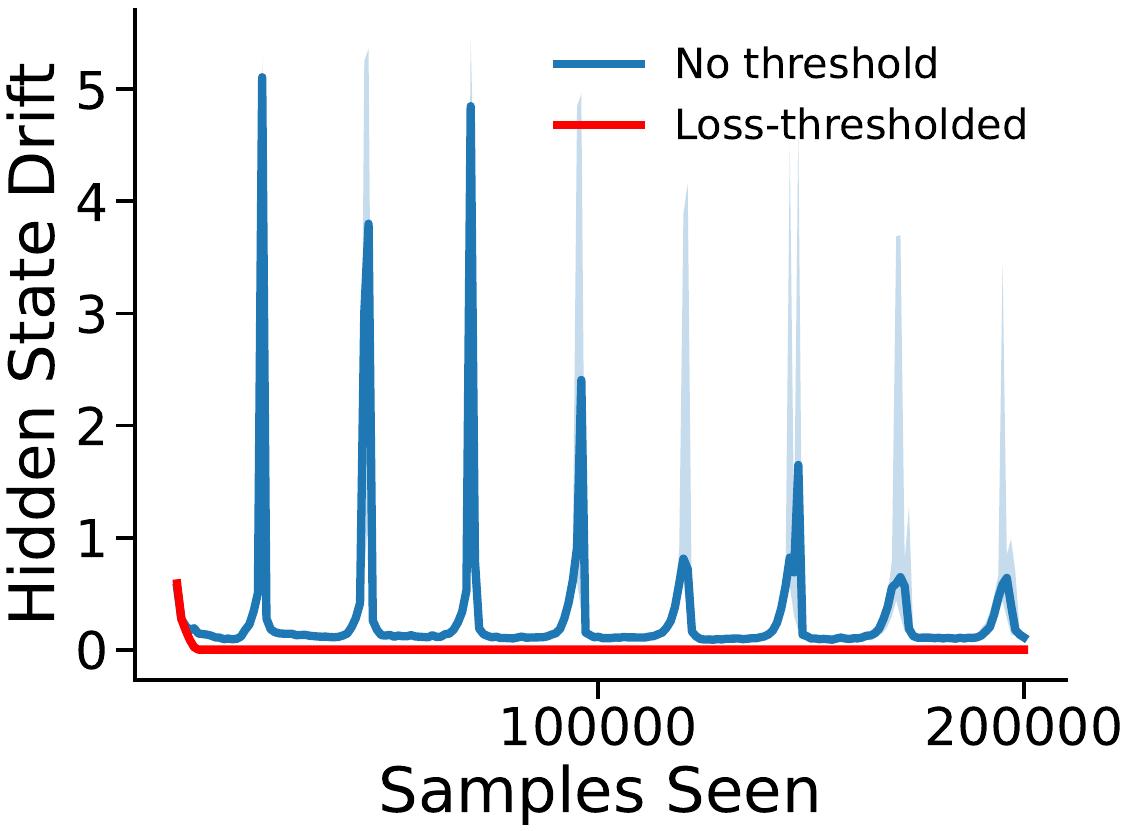}
    \caption{\textbf{Loss-thresholded updates stabilize representations, reducing drift observed under standard training.} Drift is measured as the mean $L_2$ distance between hidden states from a fixed probe sequence across checkpoints.}
    \label{fig:drift}
\end{wrapfigure}



\subsection{Experiments with Context Blocks}

\paragraph{Memory Capability Carryover across Sequential Distribution Shifts}
We train an RNN autoencoder with hidden size $100$ on one of three simulated sequence regimes using varying BPTT windows $T$, and evaluate memory retention across all regimes. Retention is measured via a linear probe that reconstructs a past token $x_i$ from the hidden state $h_t$, where $i \leq t$. Varying the offset $t - i + 1$ and measuring the reconstruction error of the probe provide a direct estimate of how much of the past information is preserved in $h_t$. Each row from left to right of Figure~\ref{fig:mem_cap} reveals intrinsic differences in regime difficulty, with the linear regime being the easiest. The off-diagonal panels show cross-distribution retention on unseen regimes. Memory representations learned on harder regimes transfer more effectively to easier ones, while representations learned on easier regimes are less reliable under harder test regimes. This indicates that the learned representations preserve sufficient temporal information under harder to easier sequence shifts.

Inspired by this observation, we quantify sequence hardness using the decoder reconstruction error and update memory weights only when this error exceeds a threshold. To reduce noise, we maintain an exponential moving average of reconstruction error with smoothing factor $0.1$. This selective update biases learning toward harder sequences while avoiding unnecessary computation on easier inputs.

\begin{wrapfigure}[14]{r}{0.38\textwidth}
    \vspace{-20pt}
    \centering
    \includegraphics[width=0.36\textwidth]{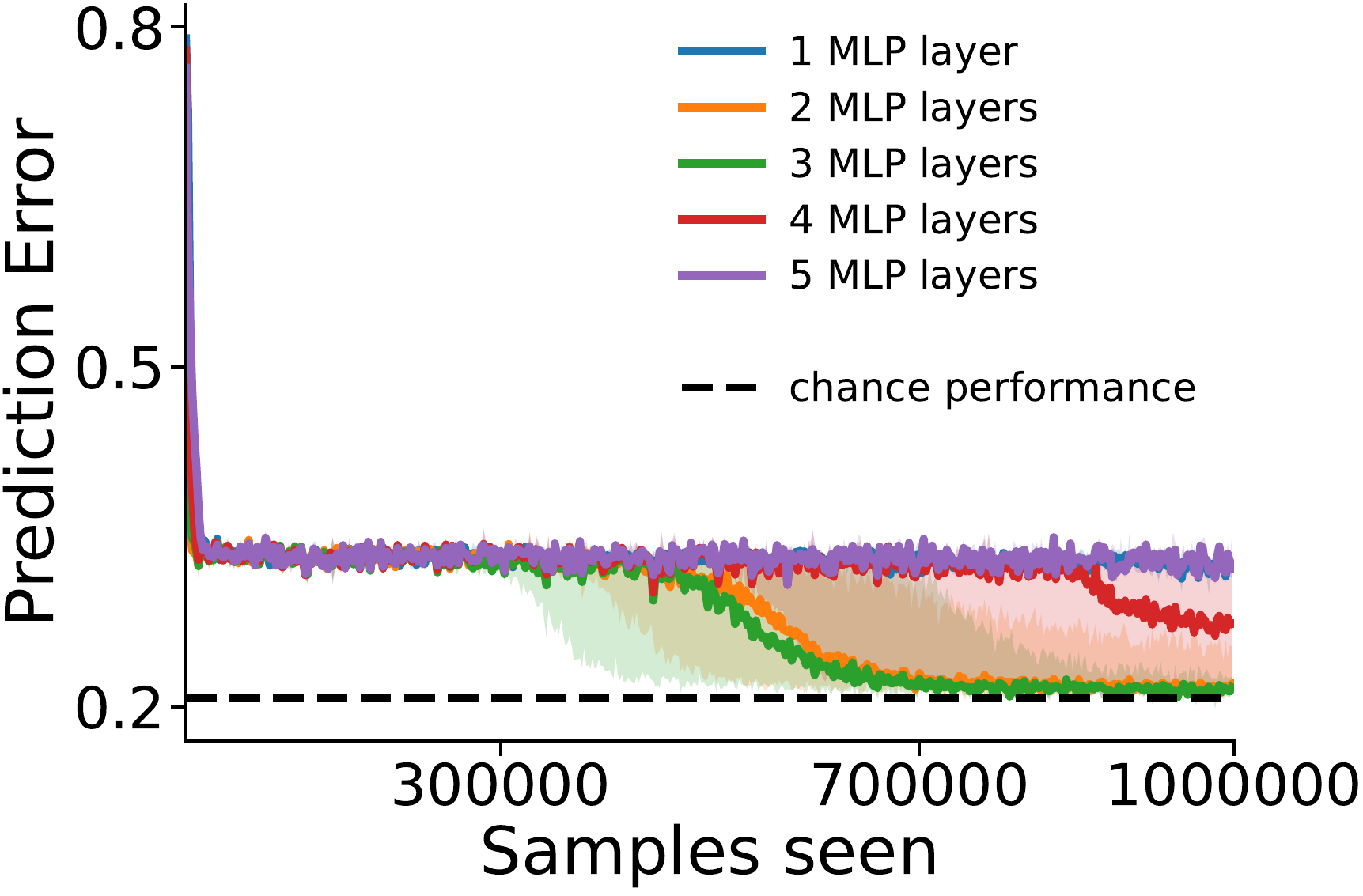}
    \caption{\textbf{Increasing the depth of the pattern recognition head improves performance up to an optimal depth, after which deeper heads slow learning.}}
    \label{fig:head}
\end{wrapfigure}

\paragraph{Hidden State Drift}
The RNN autoencoder is an overparameterized model with a non-identifiable latent space, meaning that multiple hidden state configurations can yield similar reconstruction error. As a result, the hidden representations can drift along the solution manifold during training without degrading reconstruction performance. This drift introduces a moving target for downstream pattern-recognition blocks, making it harder for them to learn stable mappings from memory states. To quantify this drift effect, we measure the mean $L_2$ distance between hidden states extracted from a fixed probe sequence at consecutive training checkpoints (every 1000 steps) on nonlinear simulation setting. Figure~\ref{fig:drift} shows standard training without any update constraint exhibits persistent representational drift. In contrast, applying a reconstruction-error-based threshold to gate memory updates mitigates this drift, leading to more stable representations once training loss stabilizes.

\subsection{Experiments with Pattern Recognition Blocks}
We study the effect of pattern recognition head complexity by varying the depth of the MLP in Figure \ref{fig:head}. As shown, increasing depth initially reduces the number of samples required to reach near-optimal performance, with the best performance achieved at depth $3$. Beyond this point, deeper heads slow down learning, indicating a trade-off between representational capacity and sample efficiency. These results suggest the existence of an optimal complexity for the pattern recognition head. For simplicity and to keep experiments computationally tractable, we fix the MLP depth to $2$ in all subsequent experiments. However, we note that further tuning of the architectural components described above may yield additional performance gains.

\begin{wrapfigure}[23]{r}{0.6\textwidth}
    \centering
    \includegraphics[width=0.58\textwidth]{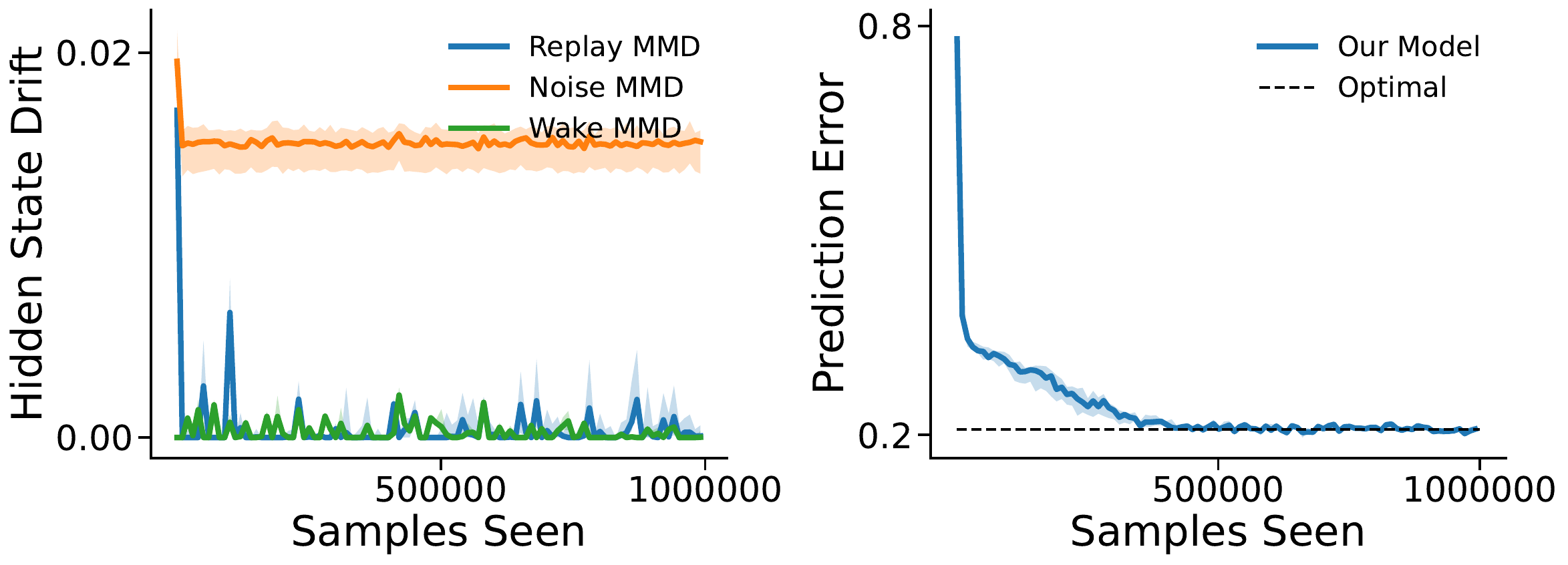}
    \caption{\textbf{Sleep replay converges to the wake hidden-state distribution while prediction performance improves.}
\textit{Left:} Distributional discrepancy between hidden states measured using Maximum Mean Discrepancy (MMD). Replay MMD (blue) compares sleep-generated states with wake states, the Wake MMD baseline (green) measures the discrepancy between two subsets of wake states, and Noise MMD (orange) compares wake states with random noise. As training progresses, replay discrepancy rapidly decreases and approaches the wake baseline while remaining far below the noise reference, indicating that replayed states become indistinguishable from wake representations.
\textit{Right:} Prediction error over training samples.}
    \label{fig:generation}
\end{wrapfigure}

\subsection{Wake Experience Generation and Sleep Replay}
The sleep phase in Figure \ref{fig:model} illustrates how wake-time experience is generated offline. At the lowest level, Pattern Block $1$ predicts the next token conditioned on the current memory state from Context $1$, $h_t^1$, and the modulating 
context from Block $2$, $c_t^2$. During the wake phase, while interacting with the environment, salient experiences, identified by reconstruction errors that exceed a fixed threshold, are tagged and stored as context pairs $(h_t^1, c_t^2)$. In this work, we maintain a fixed-size queue buffer of such context tags. Sleep phase is triggered at a regular interval. During sleep, Pattern Block $1$ generates sequences by conditioning on these stored context tags. The predicted token is fed back into Context $1$, updating its memory state $h_t^1$, which in turn influences subsequent predictions. Through this iterative process, the model can generate a particular wake-time sequence segment. Memory blocks are updated sequentially during sleep: at each step, one block is trained while all others remain frozen. This enables stable consolidation of representations across layers, preventing higher-level updates from depending on unstable lower-layer dynamics.


To quantify the quality of sleep-time generation, we measure the Maximum Mean Discrepancy (MMD) between the hidden-state distributions of wake and sleep-time at Block $1$ on nonlinear simulation. As shown in Figure \ref{fig:generation}, the discrepancy between sleep-generated and wake-time hidden-states decreases as the pattern recognition head improves.

\subsection{Ablation Study}

\begin{figure}[!ht]
    \centering
    \includegraphics[width=.8\textwidth]{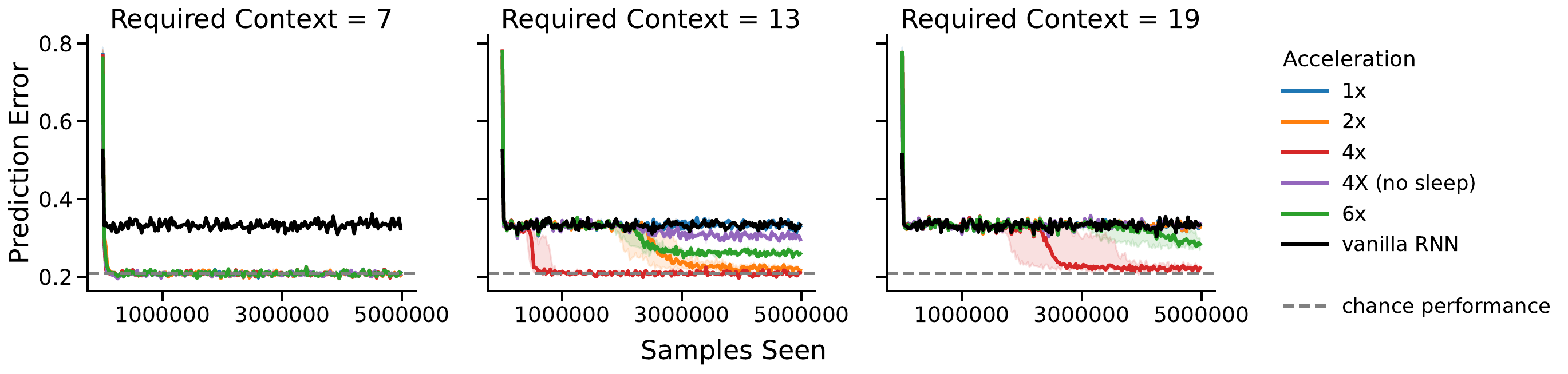}
\caption{
\textbf{Acceleration in \sct{SHARP} is constrained by memory span.}
Too little acceleration limits temporal reach, while excessive acceleration discards information. The $4\times$ no-sleep variant converges more slowly, and vanilla \sct{RNN} fails once required context exceeds the BPTT horizon.
}

    \label{fig:ablation}
\end{figure}

In this experiment, we compare a 3-layer \sct{SHARP} model with a 3-layer vanilla \sct{RNN} on the nonlinear simulation under identical constraints ($\mathrm{BPTT}=4$). We vary the required temporal context by increasing the number of past community dependencies needed to predict the next token. Figure~\ref{fig:ablation} shows that the acceleration factor must be matched to the available memory span to effectively extend the contextual reach of the pattern blocks. While larger acceleration enables access to longer temporal dependencies, excessive acceleration compresses temporal structure beyond what the memory span (set by BPTT) can preserve, leading to degraded performance. In this work, we set the acceleration factor equal to the BPTT length across all experiments. Interestingly, higher-level representations become effective only after sufficient training, leading to delayed but sharper gains for longer context requirements. This reflects an implicit curriculum over temporal scales: short-range dependencies are learned first, followed by progressively longer-range structure as hierarchical memory stabilizes.

We further include an ablation of the sleep mechanism by comparing the $4\times$ model with and without sleep. In this experiment, wake-time computation is unchanged: context states evolve according to the hierarchical downsampling schedule, the lowest context block is updated by its reconstruction objective, and pattern blocks are updated by prediction loss. The only removed component is the offline sleep phase, i.e., replay-based consolidation updates to higher context blocks. We observe that sleep does not provide uniform improvement in all regimes. Instead, its benefit emerges only when the required temporal context significantly exceeds the effective memory span of the lowest level, at which point higher-level representations become useful. When the task is solvable within the local credit assignment horizon, sleep provides no advantage.

Crucially, under the same constraint, the vanilla \sct{RNN} fails to reach optimal performance once the required context exceeds the BPTT horizon. In contrast, \sct{SHARP} successfully captures longer dependencies through accelerated replay, without increasing the length of BPTT steps. This highlights that despite having multiple layers, the vanilla \sct{RNN} fails to effectively utilize its hierarchical structure. Moreover, unlike \sct{RNN}, the context knowledge base weights in \sct{SHARP} are not updated during the active (wake) phase, further improving computational efficiency. A detailed pseudocode explaining the approach is provided in the Appendix \ref{app:pseudocode}.
\section{Benchmark Experiments}


We evaluate the proposed \sct{SHARP} model in a single-pass streaming character and subword level language modeling setting, where observations arrive sequentially without revisiting past data. This setup directly tests a model’s ability to retain and utilize long-range information under limited credit assignment, which is central to our formulation of the framework. To this end, we consider two standard character-level benchmarks with complementary properties: \texttt{text8} \citep{mahoney2011large} and \texttt{PG-19} \citep{rae2019compressive}, both of which exhibit non-stationarity in their token distributions, with \texttt{PG-19} consisting of long book-length sequences with substantial long-range dependencies (see Appendix Figure \ref{fig:nonstationarity}). Together, these datasets allow us to evaluate both short-term prediction and long-horizon memory retention. However, we report results on \texttt{PG-19} for a single run due to our computational resource limitations.

We compare against standard recurrent architectures including vanilla \sct{RNN}, \sct{GRU}, and \sct{LSTM} with identical embedding size ($100$), hidden dimension ($512$), and depth ($L=5$). All recurrent baselines are trained online with truncated context ($T=4$), with hidden states propagated sequentially. We additionally include a \sct{Clockwork~RNN} \citep{koutnik2014clockwork} with multi-timescale modules. Moreover, to contextualize performance with direct context access, we include \sct{Transformer} baselines with comparable parameter budgets. \sct{Transformer}s operate on fixed-length windows (e.g., $T=\alpha^L=1024$) without internal memory, serving as a reference for explicit context rather than streaming memory. In addition to the total parameters, we report the number of active parameters updated during the online phase, reflecting the complexity of wake-time training. Preprocessing and hyperparameter details are provided in Appendices \ref{app:data_process} and \ref{app:hyperparams}.

\paragraph{Evaluation Protocol}
Performance is measured using bits-per-character (BPC) \citep{chung2016hierarchical}, defined as $\mathbb{E}\left[-\log_2 p(x_{t+1} \mid x_{\leq t})\right]$. BPC provides a calibrated, information-theoretic measure of predictive uncertainty, making it particularly suitable for sequential settings with partial stochasticity, where accuracy alone may be misleading. BPC captures how well the model assigns probability mass to the true next token, rather than measuring prediction correctness only and lower BPC indicates better performance. We report three complementary metrics: \textbf{Forward BPC} is evaluated on unseen future data (held-out 1M tokens), \textbf{Current BPC} on the most recent 1M tokens to assess short-term adaptation, and \textbf{Backward BPC} on early training data (first 1M tokens) to measure past performance retention.

\begin{table}[!ht]
\centering
\caption{\textbf{Forward, backward, and current BPC on \texttt{text8}.} Error bars show standard deviation across $9$ runs. The Context-1 block ($\sim$1M parameters, within $5$M active) is updated only when reconstruction loss exceeds a threshold, so its updates become increasingly infrequent during training. $T$ denotes the BPTT window or input length, and $\alpha$ the acceleration factor; here, for \sct{SHARP} acceleration $\alpha=T$, total layers $L=5$. Wall-time is reported as amortized elapsed time per $1000$ online next-token predictions using identical computational resources.}
\small
\setlength{\tabcolsep}{4pt}
\resizebox{\textwidth}{!}{
\begin{tabular}{lcccccccc}
\toprule
Model & Fwd.$\downarrow$ & Cur.$\downarrow$ & Bwd.$\downarrow$ & $T$ & Time Complexity & Effective Context & Params & Wall-time / $1$k Tokens (sec) \\
\midrule
\sct{RNN} & $2.47 \pm 0.02$ & $2.41 \pm 0.02$ & $2.45 \pm 0.02$ & $4$ & $O(T)$ & $T$ & $2.4$M & $4.41$ \\
\sct{LSTM} & $2.61 \pm 0.58$ & $2.54 \pm 0.61$ & $2.50 \pm 0.54$ & $4$ & $O(T)$ & $T$ & $9.7$M & $20.50$ \\
\sct{GRU} & $2.41 \pm 0.03$ & $2.34 \pm 0.03$ & $2.38 \pm 0.03$ & $4$ & $O(T)$ & $T$ & $7.3$M & $14.29$ \\
\sct{Clockwork~RNN} & $2.95 \pm 0.06$ & $2.79 \pm 0.06$ & $2.92 \pm 0.07$ & $4$ & $O(T)$ & $T$ & $4.3$M & $4.91$ \\
\midrule
\sct{Transformer~5M} & $2.41 \pm 0.02$ & $2.33 \pm 0.04$ & $2.38 \pm 0.02$ & $1024$ & $O(T^2)$ & $T$ & $5.0$M & $116.40$ \\
\sct{Transformer~10M} & $2.43 \pm 0.03$ & $2.37 \pm 0.04$ & $2.41 \pm 0.03$ & $1024$ & $O(T^2)$ & $T$ & $9.7$M & $266.53$ \\
\midrule
\sct{SHARP} & $\mathbf{2.32} \pm 0.06$ & $\mathbf{2.23} \pm 0.07$ & $\mathbf{2.30} \pm 0.06$ & $4$ & $O(T)$ & $\alpha^L$ & $9.7$M ($5$M active) & $9.46$\tablefootnote{\footnotesize For \sct{SHARP}, $9.46$s is amortized per $1000$ online tokens: wake cost is $7.31$s/$1$k tokens, and sleep adds $43.03 \times (1000/20000)=2.15$s because sleep is triggered once every $20{,}000$ wake steps; thus $7.31+2.15=9.46$s.} \\
\bottomrule
\end{tabular}
}
\label{tab:text8}
\end{table}

\begin{figure}[!ht]
    \centering
    \includegraphics[width=.8\textwidth]{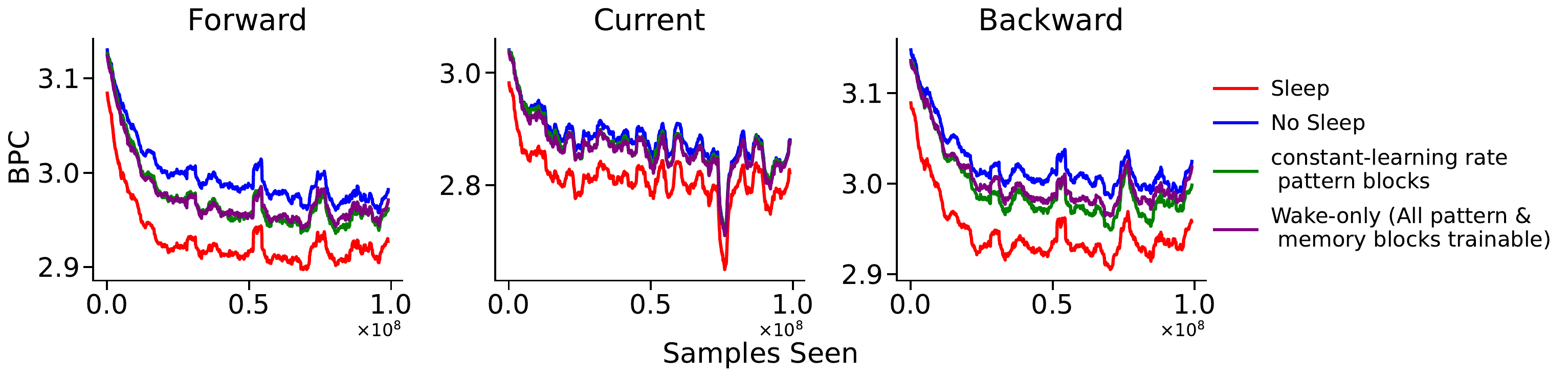}
\caption{\textbf{Sleep ablations on \texttt{text8}.} Sleep-enabled \sct{SHARP} achieves the lowest forward, current, and backward BPC.}

    \label{fig:ablation_text8}
\end{figure}

\begin{figure}[!ht]
    \centering
    \includegraphics[width=.85\textwidth]{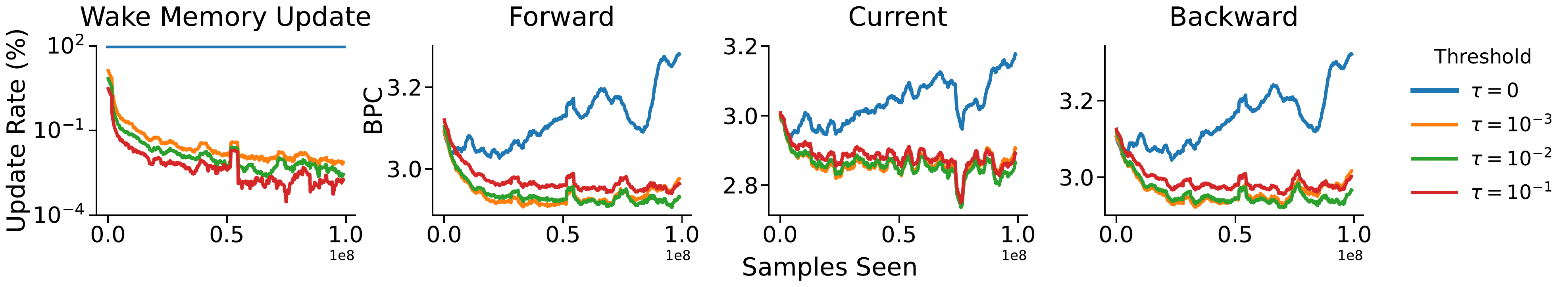}
\caption{\textbf{Thresholded updates improve computational efficiency and performance stability.} Performance is insensitive to the choice of smaller thresholds, while removing thresholding causes performance to degrade over time.}
    \label{fig:threshold_text8}
\end{figure}

\subsection{Character-Level Modeling}
\paragraph{Results on \texttt{text8}}


Table~\ref{tab:text8} shows that on \texttt{text8}, \sct{SHARP} consistently achieves the best performance across forward, current, and backward BPC, indicating improved retention, quick adaptation, and generalization in the streaming setting. Moreover, the performance of the baselines reflects their architectural specifications. Vanilla \sct{RNN}s perform competitively due to their simplicity and stable optimization under short BPTT, but lack mechanisms to extend effective context beyond the truncation horizon. \sct{LSTM}s exhibit high variance across runs, likely due to the difficulty of reliably gating long-range dependencies under strict single-pass training and limited credit assignment, leading to unstable memory retention. \sct{GRU}s provide a more stable trade-off between expressivity and optimization, resulting in relatively consistent performance but still constrained by short-term credit assignment. \sct{Clockwork~RNN}, despite its multi-timescale design, operates through a horizontal hierarchy of modules that differ only in their update frequency. While this enables modeling of multiple temporal resolutions, it does not induce a hierarchy of representations: all modules operate within the same representational space without progressively transforming or abstracting information. As a result, there is no notion of higher-level generalized representations or lower-level specialized features. In contrast, \sct{SHARP} constructs a vertical hierarchy in which representations are progressively compressed and reorganized across layers, enabling structured abstraction of temporal information. Transformer models achieve strong performance by directly attending to recent inputs, but rely on explicit access to past tokens and incur quadratic complexity, making them less suitable for strict streaming constraints. In contrast, \sct{SHARP} separates memory from prediction and propagates information hierarchically via acceleration, enabling efficient long-range retention with linear complexity. 

We further ablate \sct{SHARP} on \texttt{text8} by considering three variants: removing the offline sleep phase, removing the slower learning-rate schedule for upper pattern blocks, and training the full model during wake only without freezing the context knowledge base. For faster experimentation, these ablations use a smaller hidden size of $128$. Figure~\ref{fig:ablation_text8} shows that the sleep-enabled model consistently achieves the lowest BPC across forward, current, and backward evaluations. Removing sleep degrades performance, indicating that offline replay-based consolidation improves future generalization, current adaptation, and retention of earlier stream statistics. The no-pattern-slowdown and wake-only all-trainable variants also underperform the full sleep-enabled model, suggesting that both sleep-time consolidation and hierarchical timescale separation contribute to \sct{SHARP}'s gains.

We also experiment on the sensitivity of performance depending on the update threshold. Figure~\ref{fig:threshold_text8} shows that reconstruction-thresholded memory updates rapidly reduce wake-time context-block updates to near zero after the initial training phase. Moderate thresholds ($\tau=10^{-3},10^{-2}$) achieve the best forward, current, and backward BPC, while updating at every step ($\tau=0$) causes performance to degrade over time. This suggests that thresholding both reduces computation and prevents unnecessary memory drift from over-updating on already well-reconstructed inputs.

\begin{table}[!h]
\centering
\scriptsize
\setlength{\tabcolsep}{2.5pt}
\renewcommand{\arraystretch}{0.92}

\caption{\textbf{PG-19 benchmark results.} Left: character-level BPC on PG-19. Right: subword-level BPT using pretrained \sct{GPT-2} tokenizer and frozen embeddings. All results are from a single run.}
\label{tab:pg19_combined}

\begin{minipage}[t]{0.48\textwidth}
\centering
\textbf{(a) Character-level PG-19}
\vspace{0.5ex}

\begin{tabular}{lcccc}
\toprule
Model & Fwd.$\downarrow$ & Cur.$\downarrow$ & Bwd.$\downarrow$ & $T$\\
\midrule
\sct{RNN} & $2.79$ & $2.86$ & $2.94$ & $4$\\
\sct{LSTM} & $3.44$ & $3.36$ & $3.55$ & $4$\\
\sct{GRU} & $2.43$ & $2.48$ & $2.48$ & $4$\\
\sct{Clockwork RNN} & $2.46$ & $2.51$ & $2.70$ & $4$\\
\midrule
\sct{Transformer 10M} & $2.13$ & $2.16$ & $2.24$ & $1024$ \\
\sct{Transformer 5M} & $2.15$ & $2.18$ & $2.26$ & $1024$ \\
\midrule
\sct{SHARP} & $2.29$ & $2.35$ & $2.40$ & $4$\\
\bottomrule
\end{tabular}
\end{minipage}
\hfill
\begin{minipage}[t]{0.48\textwidth}
\centering
\textbf{(b) Subword-level PG-19}
\vspace{0.5ex}

\begin{tabular}{lcccc}
\toprule
Model & Fwd.$\downarrow$ & Cur.$\downarrow$ & Bwd.$\downarrow$ & $T$\\
\midrule
\sct{RNN} & $4.54$ & $4.56$ & $4.55$ & $4$\\
\sct{LSTM} & $4.52$ & $4.54$ & $4.53$ & $4$\\
\sct{GRU} & $4.50$ & $4.52$ & $4.51$ & $4$\\
\midrule
\sct{Transformer-18M} & $3.95$ & $3.77$ & $3.94$ & $256$\\
\sct{Transformer-23M} & $3.94$ & $3.76$ & $3.94$ & $256$\\
\midrule
\sct{SHARP} & $4.26$ & $4.26$ & $4.26$ & $4$\\
\bottomrule
\end{tabular}
\end{minipage}

\end{table}

\paragraph{Results on \texttt{PG-19}}
\texttt{PG-19} introduces stronger distribution shifts across books. Consistent with \texttt{text8}, \sct{SHARP} demonstrates improved forward generalization and reduced backward degradation, suggesting that hierarchical replay enables better transfer of learned structure across distributions (Table \ref{tab:pg19_combined}a). In contrast, standard recurrent baselines exhibit significant performance gaps between current and backward metrics, indicating instability under distribution shift. Compared with the character-level setting in \texttt{text8}, the gap to Transformers is larger because Transformers directly attend over a $1024$-token window, whereas \sct{SHARP} uses only local wake-time credit assignment. Moreover, because Transformers have direct access to the input window, they rely less on internal memory formation and can devote more capacity to predictive mapping over the visible context.

\subsection{Subword-Level Modeling on \texttt{PG-19}}
We further evaluate whether \sct{SHARP} can utilize pretrained representations in a more realistic subword-level setting with a $50{,}257$-token \sct{GPT-2} vocabulary, rather than the $27$-character vocabulary used above. In contrast to the character-level setup, where text is lowercased and punctuation is removed, the subword setup preserves richer lexical structure through the \sct{GPT-2} tokenizer. We use pretrained, frozen \sct{GPT-2} token embeddings as the input front end for both \sct{SHARP} and recurrent baselines, and train the models online to predict the next subword token under the same strict streaming constraint with BPTT $=4$. For faster runtime, we use $4$ SHARP hierarchy levels and $4$ MLP layers per pattern block while keeping other settings unchanged from character-level modeling. We trained on the first $217$ books without any truncation. As shown in Table~\ref{tab:pg19_combined}b, all models achieve BPT far below the chance level of $\log_2(50{,}257)\approx 15.62$, and \sct{SHARP} improves over recurrent baselines across forward, current, and backward BPT. \sct{SHARP}'s consistent improvement over \sct{RNN}, \sct{LSTM}, and \sct{GRU} suggests that hierarchical accelerated replay remains useful beyond low-level character transitions, extending to larger subword vocabularies with pretrained semantic embeddings.

\section{Discussion}

We introduced \sct{SHARP}, a hierarchical framework that separates memory (without credit-assignment) from pattern recognition (with credit-assignment), extending the temporal context via accelerated replay of lower-level memory. Our experiments suggest that separating memory from prediction enables \sct{SHARP} to extend effective context beyond the BPTT horizon through progressively compressed temporal summaries. This produces a hierarchical organization in which higher levels capture longer-range dependencies without requiring long-horizon wake-time credit assignment.


While our current instantiation demonstrates promising results, it can be optimized further. The framework is modular and agnostic to the choice of memory representation: more expressive memory modules (e.g., \sct{LSTM} autoencoders) or alternative mechanisms may further enhance acceleration and extend the effective temporal horizon. Similarly, improving pattern recognition modules to better utilize the available context may further enhance performance. In addition, \sct{SHARP} can be integrated with modern state space models such as \sct{Mamba} to construct temporal hierarchies through acceleration, enabling efficient long-range sequence modeling. 

In future work, we aim to explore alternative lowest-level episodic memory mechanisms, such as Hebbian learning inspired by hippocampal place cells, where memory is represented as a superposition of basis elements. Each basis corresponds to a token or a contiguous input segment. Such episodic memory representations, where each memory element can be identified separately, may enable downstream pattern modules to perform selective retrieval akin to attention. This suggests a pathway toward attention-like mechanisms in streaming settings, where relevance emerges during retrieval rather than being predetermined at memory encoding. Moreover, the geometry of learned semantic memory embeddings (see Equation \ref{eq:mem_property}) at the upper levels, where individual memory elements are not separately identifiable, may serve as an additional substrate for storing long-term memory. This would complement the short-term memory mechanisms studied in this paper while further extending the effective context reach of the system. More broadly, we emphasize that not only the capacity but also the quality of memory representations should be studied in depth along with pattern-recognition mechanisms to build more robust and generalizable continual learning systems.


\section{Acknowledgment}
This work was graciously supported by the NSF EFRI under Award $2317706$ and was supported in part by NSF NAIAD under award 2332744, and the U.S. Department of Energy's (DoE) ESTEEM Center. 

\paragraph{\textbf{Author Contributions:}} J.D. conceived the SHARP framework, including the sleep-based hierarchical accelerated replay mechanism, memory/prediction separation, model architecture, and algorithmic design. J.D. implemented the model, ran the primary experiments, analyzed the results, prepared the figures, drafted the manuscript, and led the revision and rebuttal preparation. S.S. ran selected baseline experiments, including Transformer and selected recurrent-model comparisons. I.L. provided feedback on the neuroscience framing and interpretation of sleep-dependent temporal learning. C.K. provided feedback on the machine learning framing and evaluation. D.K. provided grant supervision, funding support, and feedback on the manuscript.

\bibliography{ref}
\bibliographystyle{collas2026_conference}

\clearpage
\appendix
\section{Simulation Environments}
\label{app:sim}

 \paragraph{Linear:} In this simulation, the token sequence \{A, B, C, D, E, F, G\} is repeated periodically. We refer to this setting as \emph{Linear}, since a sequence model with purely linear dynamics should, in principle, be able to capture the underlying periodic rule without requiring nonlinear transformations.

\begin{wrapfigure}[16]{r}{0.41\textwidth}
    \centering
    \includegraphics[width=0.39\textwidth]{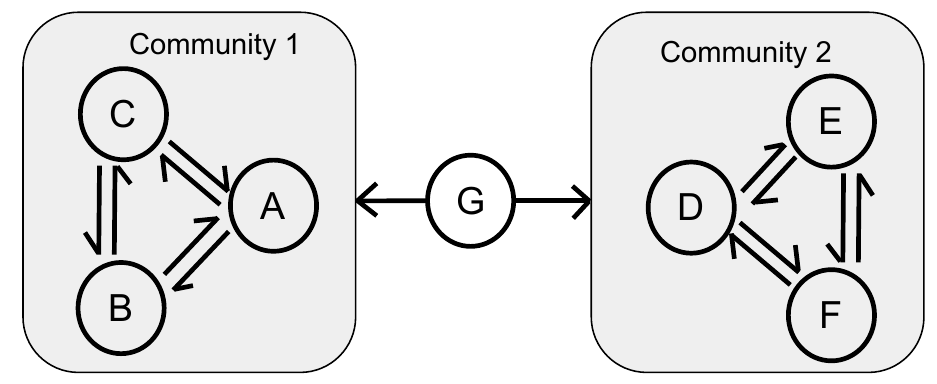}
    \caption{\textbf{Token generation graph for the nonlinear simulation.}
Two communities, Community 1 \{A, B, C\} and Community 2 \{D, E, F\}, are connected by a hub token G. From G either community can be entered with equal probability. The traversal direction depends on the past $k$ community visits.}
    \label{fig:simulation}
\end{wrapfigure}

\paragraph{Nonlinear:}
Figure~\ref{fig:simulation} illustrates the transition graph for the nonlinear simulation. The system consists of two token communities: Community~1 \{A, B, C\} and Community~2 \{D, E, F\}. Within each community, tokens are traversed in either a clockwise (ABC, BCA, CAB, DEF, EFD, FDE) or counterclockwise (ACB, CBA, BAC, DFE, FED, EDF) direction. A special token G acts as a hub connecting the two communities. From G, the next token is sampled uniformly from \{A, B, C, D, E, F\}, thereby selecting both the community and the starting point of traversal. Once a community is entered, the model completes exactly one full traversal (three steps) before deterministically returning to G.

Crucially, the traversal direction is determined by the parity of the last $K$ community visits. Let $v_{t-K}, \ldots, v_{t-1} \in \{0,1\}$ denote the community indices of the previous $K$ visits. If $\sum_{i=t-K}^{t-1} v_i$ is even, traversal proceeds clockwise; otherwise, it proceeds counterclockwise. Thus, correct prediction requires retaining memory over past $K$ visits, inducing a fixed-length temporal dependency. An example sequence generated for $K=2$ is: `CAB-G-DEF-G-DFE-G-ABC-G-FED-G-EDF-G-FED-G-FDE-G-DEF-G-EFD-G-EFD'. Note that for $K=2$, the direction at the second token of a community depends on the preceding $7$ tokens.

The environment is partially stochastic: while traversal within a community is deterministic given the direction, the transition from $G$ is uniform over six tokens. Consequently, the optimal achievable accuracy is $\frac{3 + \frac{1}{6}}{4} = 79.17\%$.

\paragraph{Random:} This simulation represents the theoretically most difficult case to retain. At each time step, a token is sampled uniformly at random from the set \{A, B, C, D, E, F, G\}. Since the sequence contains no underlying structure, a model cannot utilize implicit regularities to compress its memory footprint. 

\newpage
\section{Pseudocode}
\label{app:pseudocode}

\begin{algorithm}[!htbp]
\caption{\sct{SHARP}: Wake and Sleep Phases}
\label{alg:sharp}
\begin{algorithmic}[1]
\Require input stream $\{x_t\}$, context blocks $\{m^\ell\}_{\ell=1}^{L}$, pattern blocks $\{f^\ell\}_{\ell=1}^{L}$, acceleration factor $\alpha$

\State initialize memory states $\{h^\ell\}_{\ell=1}^{L}$ and context queue buffer $\mathcal{B}$

\For{each time step $t=1, 2, \cdots$}

    \Comment{\textbf{Wake phase}}

    \State update lowest memory:
    \[
    h^1_t \leftarrow m^1(x_t, h^1_{t-1})
    \]

    \State \textbf{(selective update)} update $m^1$ using reconstruction loss if needed

    \State propagate states bottom-up with acceleration:
    \[
    h^\ell_t \leftarrow m^\ell(h^{\ell-1}_t, h^\ell_{t-1}) \quad \text{if } t \equiv 0 \pmod{\alpha^{\ell-1}}, \text{ for } \ell = 2, \cdots, L
    \]

    \State construct top-down context:
    \[
    c^\ell_t \leftarrow f^\ell(h^\ell_t, c^{\ell+1}_t), \text{for}~\ell = L, \cdots, 2
    \]

    \State predict next state transition probability:
    \[
    \hat{\mathbf{p}}_t \leftarrow f^1(h^1_t, c^2_t)
    \]

    \State update pattern-recognition blocks using prediction loss

    \If{memory update is triggered}
        \State store $(h^1_t, c^2_t)$ in buffer $\mathcal{B}$ \Comment{context tags}
    \EndIf

~
    \If{sleep is triggered}

        \Comment{\textbf{Sleep phase}}
        \For{each layer $\ell = 2, \dots, L$}
            \State sample tagged context from buffer $\mathcal{B}$
            \State replay $(l-1)$th layer states with stride $\alpha$ 
            \State update $m^\ell$ to reconstruct replayed states
        \EndFor

    \EndIf

\EndFor
\end{algorithmic}
\end{algorithm}
\section{Data Preprocessing and Evaluation Protocol}
\label{app:data_process}
\begin{table}[t]
\centering
\caption{
Hyperparameters and configuration choices used by \sct{SHARP}. 
We distinguish method-specific hyperparameters from standard architecture, data, and optimization settings.
}
\label{tab:sharp_hyperparameters}
\small
\begin{tabular}{lll}
\toprule
\textbf{Category} & \textbf{Hyperparameter / Setting} & \textbf{Role} \\
\midrule
\multicolumn{3}{l}{\textit{Method-specific hyperparameters}} \\
\midrule
Hierarchy & Number of layers $L$ & Number of context and pattern-recognition levels \\
Temporal scaling & Acceleration factor $\alpha$ & Downsampling factor between adjacent context levels \\
Pattern slow-down & learning rate slow-down factor $\gamma$ & updates upper pattern blocks more slowly\\
Memory update & Reconstruction threshold $\tau$ & Triggers selective memory updates and context tagging \\
Replay storage & Context buffer size & Number of tagged wake states retained for sleep replay \\
Sleep schedule & Sleep interval & Frequency of offline consolidation phases \\
Replay length & Replay sequence length & Number of replayed states used during sleep updates \\
\midrule
\multicolumn{3}{l}{\textit{Architecture and scale settings}} \\
\midrule
Pattern block & MLP depth & Number of layers in each pattern-recognition head \\
Representation & Hidden size & Dimensionality of memory and recurrent states \\
Input encoding & Embedding dimension & Dimensionality of token embeddings \\
Data-dependent & Vocabulary size & Number of discrete tokens in the dataset \\
\midrule
\multicolumn{3}{l}{\textit{Training and optimization settings}} \\
\midrule
Credit horizon & BPTT window $T$ & Local wake-time credit-assignment horizon \\
Optimization & Optimizer & Algorithm used for gradient-based updates \\
Optimization & Learning rate & Step size for trainable parameters \\
Regularization & Weight decay & Weight regularization coefficient \\
\bottomrule
\end{tabular}
\end{table}

\subsection{Text8}
We use the standard \texttt{text8} corpus as a continuous character stream. A character-level vocabulary is constructed directly from the dataset, and the text is encoded into integer token IDs. To obtain error bar estimates under a single-pass setting, we partition the corpus into $9$ disjoint segments of $10$M characters each, and train a separate model on each segment independently. Training within each segment is performed sequentially without shuffling.

Evaluation is performed within each segment using three disjoint portions of the stream: the initial $1$M tokens are used to assess retention (backward), the most recent $1$M tokens of the training stream are used for current performance, and a held-out future portion (last $1$M tokens of \texttt{text8}) is used to measure generalization (forward). For fairness, context length at evaluation is fixed across all approaches, and the predicted (last) positions' loss is measured. Results are aggregated across the $9$ runs.

\begin{figure}[!t]
    \centering
    \includegraphics[width=.45\textwidth]{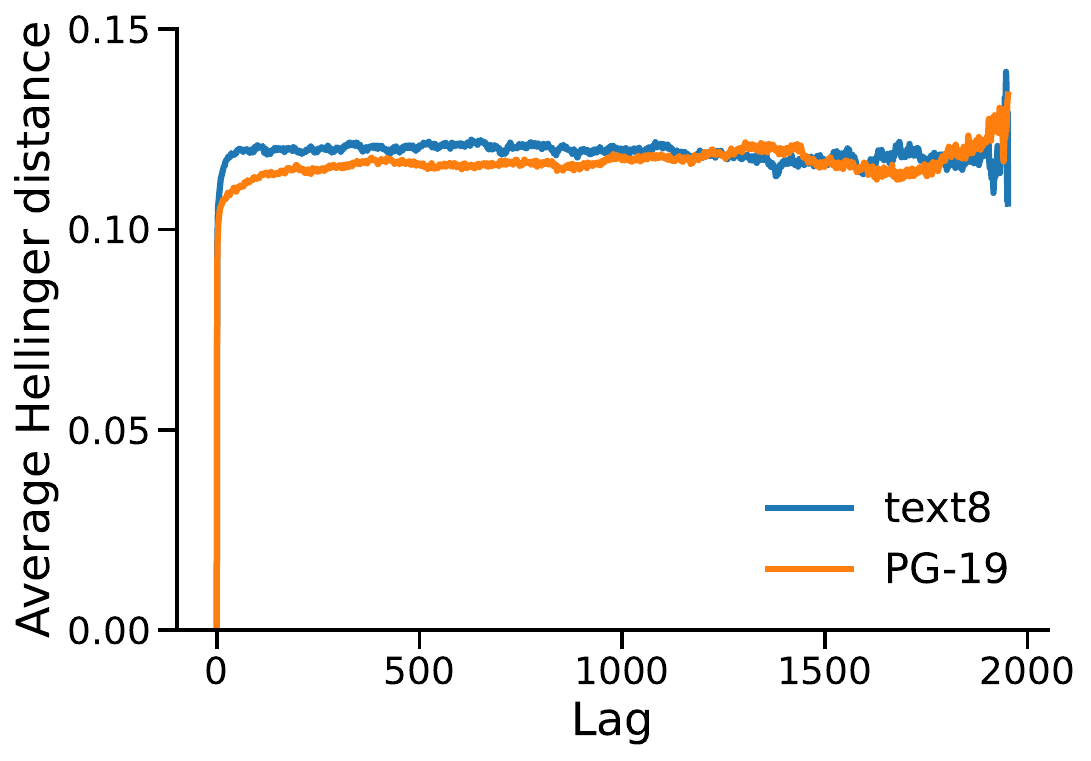}
\caption{
\textbf{Context-scale non-stationarity in \texttt{text8} and \texttt{PG-19}.}
Average Hellinger distance between character distributions (histograms) computed over non-overlapping $4^5 = 1024$-token windows, as a function of lag (window separation). Each point represents the mean distance between all window pairs separated by a fixed lag. Both datasets exhibit a sharp increase in distance at small lags followed by a plateau, indicating that local token distributions diverge rapidly but stabilize beyond a characteristic separation.
}

    \label{fig:nonstationarity}
\end{figure}

To assess non-stationarity at the scale relevant to the model, we compute the Hellinger distance between character distributions (histograms) over non-overlapping windows of length $1024$, matching the effective context size. Distances are averaged over all window pairs separated by a fixed lag. As shown in Figure~\ref{fig:nonstationarity}, the distance increases rapidly at small lags and saturates thereafter, indicating that local token distributions evolve over short temporal scales and this distribution shift reaches its maximum after a certain lag. This demonstrates that the input stream is non-stationary at the scale of the model’s available context.

\paragraph{Text8 100M Character Sequence Modeling}
\label{app:text8_100M}

To complement the nine-fold cross-validation results in the main text, we evaluate each model on the full 90\,M-character \texttt{text8} training split as a single continuous sequence, keeping the same
model configurations as our main experiments. The final 10\,M characters are held out as a test
set. This regime more closely mirrors deployment conditions where a model encounters one long,
non-repeating stream of data, and it amplifies differences in catastrophic forgetting: a model
that loses information about early portions of the stream will exhibit a large gap between its
forward (future) and backward (past) BPC.

\begin{table}[htbp]
\centering
\caption{\textbf{Forward, backward, and current performance on \texttt{text8} (100M regime, single run).}
All models trained on the full 90\,M-character text8 training split; only one run per model so no
error bars are reported. $T$ denotes the BPTT window or input length; for \sct{SHARP} $\alpha$ is
the acceleration factor. \textsuperscript{$\dagger$}~RNN backward/current BPC approaches the
random-predictor ceiling ($\log_2 27 \approx 4.76$\,BPC), indicating severe catastrophic
forgetting under 90\,M-step online training; gated architectures (LSTM, GRU) do not exhibit
this collapse.}
\small
\setlength{\tabcolsep}{4pt}
\resizebox{\textwidth}{!}{
\begin{tabular}{lccccccc}
\toprule
Model & Forward BPC $\downarrow$ & Current BPC $\downarrow$ & Backward BPC $\downarrow$ & $T$ & Time Complexity & Effective Context & Params \\
\midrule
\sct{RNN}\textsuperscript{$\dagger$}
    & $2.70$ & $4.63$ & $4.63$ & $4$ & $O(T)$ & $T$ & $2.4$M \\
\sct{LSTM}
    & $3.17$ & $3.16$ & $3.14$ & $4$ & $O(T)$ & $T$ & $9.7$M \\
\sct{GRU}
    & $2.57$ & $2.50$ & $2.56$& $4$ & $O(T)$ & $T$ & $7.3$M \\
\sct{Clockwork~RNN} & $2.88$ & $2.85$ & $2.78$ & $4$ & $O(T)$ & $T$ & $4.3$M \\
\midrule
\sct{Transformer} (ctx=$1024$)
    & $2.22$ & $2.15$ & $2.20$ & $1024$ & $O(T^2)$ & $T$ & $5.0$M \\
\sct{Transformer} (ctx=$1024$)
    & $2.19$ & $2.12$ & $2.17$ & $1024$ & $O(T^2)$ & $T$ & $9.7$M \\
\midrule
\sct{SHARP}
    & $2.36$    & $2.32$    & $2.33$    & $4$ & $O(T)$   & $\alpha^L$ & $9.7$M ($5$M active) \\
\bottomrule
\end{tabular}
}
\label{tab:text8_100M}
\end{table}

\subsection{PG-19}
For \texttt{PG-19}, all books are normalized to a fixed 27-character vocabulary consisting of \texttt{a}--\texttt{z} and space. Specifically, text is lowercased, all non-alphabetic characters are mapped to spaces, and consecutive spaces are collapsed. Training data is constructed by sequentially concatenating books from the training split until a total budget of $100$M characters is reached. Books shorter than $20$K normalized characters are discarded to avoid degenerate sequences. To ensure computational tractability and balanced contribution across books, each training book is truncated to at most $2$M characters. For evaluation, we select up to $5$ held-out books from the validation split (or test split if needed), each truncated to at most $1$M characters. These books are never seen during training and are used to assess generalization. Evaluation is performed sequentially within each book without resetting hidden states. For fairness, context length at evaluation is fixed across all approaches, and the predicted (last) positions' loss is measured. In addition, retention is evaluated on the first $3$ training books (each truncated to at most $1$M characters), while current performance is measured on the last $3$ training books under the same truncation. This allows us to characterize both memory retention and adaptation within the training stream.

To examine non-stationarity at the scale relevant to the model, we compute the Hellinger distance between character distributions (histograms) over non-overlapping windows of length $1024$ within the sequence. Distances are averaged over all window pairs separated by a fixed lag. As shown in Figure~\ref{fig:nonstationarity}, the distance increases rapidly at small lags and plateaus thereafter, indicating that local token distributions change over short temporal scales while remaining stable beyond a characteristic separation. This shows that the input stream is not locally stationary even at the scale of the model’s effective context.

\paragraph{Sequence construction}
For both datasets on recurrent baselines, the next-token prediction is formulated using a fixed context window of length $4$. Given a token sequence $\{x_t\}$, each training or evaluation sample is constructed as
\[
(x_{t-4}, x_{t-3}, x_{t-2}, x_{t-1}) \rightarrow x_{t}.
\]
These subsequences are extracted in a sliding-window fashion with stride $1$, ensuring that every position in the sequence contributes a training example.

Importantly, while inputs are constructed from short local windows, hidden states are propagated sequentially across the stream without resetting, allowing the model to accumulate information over long temporal horizons beyond the fixed context window. In contrast, Transformers do not maintain persistent hidden states across the stream and rely solely on explicit context windows, highlighting the distinction between learned memory and direct context access.

\section{Hyperparameters}
\label{app:hyperparams}

\begin{table}[htbp]
\centering
\caption{\textbf{Hyperparameters for \sct{SHARP} on Benchmark Datasets.}}
\small
\setlength{\tabcolsep}{6pt}
\begin{tabular}{lc}
\toprule
\textbf{Hyperparameter} & \textbf{Value} \\
\midrule
Total layers ($L$) & 5 \\
Pattern Block MLP layers & 2 \\
Hidden sizes & [512, 512, 512, 512, 512] \\
Embedding dimension & 100 \\
Vocabulary size & 27 \\
BPTT ($T$) & 4 \\
Acceleration factor ($\alpha$) & $4$ \\
Reconstruction threshold ($\tau$) & $10^{-2}$ \\
Learning rate & $10^{-4}$ \\
Optimizer & Adam \\
Weight decay & $10^{-12}$ \\
Context buffer size & $20$ \\
Go to sleep & after every 20000th steps\\
Replay sequence length & $1025$\\
\bottomrule
\end{tabular}
\label{tab:hyperparams}
\end{table}

\begin{table}[htbp]
\centering
\caption{\textbf{Hyperparameters for RNN, LSTM, and GRU baselines on Benchmark Datasets.}}
\small
\setlength{\tabcolsep}{6pt}
\begin{tabular}{lc}
\toprule
\textbf{Hyperparameter} & \textbf{Value} \\
\midrule
Total layers & 5 \\
Hidden size & 512 \\
Embedding dimension & 100 \\
Vocabulary size & 27 \\
BPTT & 4 \\
Learning rate & $10^{-4}$ \\
Optimizer & Adam \\
Weight decay & $10^{-12}$ \\
\bottomrule
\end{tabular}
\label{tab:rnn_hyperparams}
\end{table}

\begin{table}[htbp]
\centering
\caption{\textbf{Hyperparameters for Clockwork RNN baseline on Benchmark Datasets.}}
\small
\setlength{\tabcolsep}{6pt}
\begin{tabular}{lc}
\toprule
\textbf{Hyperparameter} & \textbf{Value} \\
\midrule
Number of modules & 5 \\
Hidden size per module & 512 \\
Total hidden size & 2560 \\
Periods & [1, 2, 4, 8, 16] \\
Embedding dimension & 100 \\
Vocabulary size & 27 \\
BPTT & 4 \\
Learning rate & $10^{-4}$ \\
Optimizer & Adam \\
Weight decay & $10^{-12}$ \\
\bottomrule
\end{tabular}
\label{tab:clockwork_hyperparams}
\end{table}

\begin{table}[htbp]
\centering
\caption{\textbf{Hyperparameters for the Transformer baseline on benchmark datasets.} Two parameter budgets ($\sim$10M, $\sim$5M for char-level tasks; $\sim$22.7M, $\sim$18.0M at for sub-word task) each use a training context length of $1024$. Architecture follows a Pre-LN LLaMA-style stack (RMSNorm, RoPE, SwiGLU). Our implementation was built on codes from~\citet{sun2025curse}.}
\small
\setlength{\tabcolsep}{6pt}
\begin{tabular}{lc}
\toprule
\textbf{Hyperparameter} & \textbf{Value} \\
\midrule
\multicolumn{2}{l}{\textit{Shared across all variants}} \\
Vocabulary size & $27$ (character-level) / $50{,}257$ (sub-word, GPT-2 BPE) \\
Input token embedding & learned, $\mathcal{N}(0, 0.02)$ (char-level) / frozen GPT-2 + linear proj.\ to $d_{\mathrm{model}}$ (sub-word) \\
Model dimension $d_{\mathrm{model}}$ & $256$ \\
Number of attention heads & $8$ (head dim.\ $32$) \\
Attention & Causal self-attention (scaled dot-product) \\
Normalization & RMSNorm (pre-attention / pre-MLP), $\varepsilon=10^{-6}$ \\
Position encoding & RoPE, base $=10^{4}$ \\
MLP & SwiGLU ($d_{\mathrm{ff}}$ as below) \\
Linear / embedding init.\ & $\mathcal{N}(0, 0.02)$ \\
Optimizer & Adam \\
Learning rate & $10^{-4}$ \\
Weight decay & $10^{-12}$ \\
Batch size & $1$ \\
\midrule
\multicolumn{2}{l}{\textit{$\sim$10M parameters (char-level) / $\sim$22.7M parameters (sub-word)}} \\
Number of layers & $12$ \\
Feedforward dim.\ $d_{\mathrm{ff}}$ & $704$ \\
Training context length $T$ & $1024$ (char-level) / $256$ (sub-word); contiguous non-overlapping chunks \\
Approx.\ parameters & $\sim 9.65\times 10^{6}$ (char-level) / $\sim 2.27\times 10^{7}$ (sub-word) \\
\midrule
\multicolumn{2}{l}{\textit{$\sim$5M parameters (char-level) / $\sim$18.0M parameters (sub-word)}} \\
Number of layers & $6$ \\
Feedforward dim.\ $d_{\mathrm{ff}}$ & $736$ \\
Training context length $T$ & $1024$ (char-level) / $256$ (sub-word); contiguous non-overlapping chunks \\
Approx.\ parameters & $\sim 4.98\times 10^{6}$ (char-level) / $\sim 1.80\times 10^{7}$ (sub-word) \\
\bottomrule
\end{tabular}
\label{tab:transformer_hyperparams}
\end{table}

\end{document}